\documentclass[runningheads]{llncs}

\usepackage{graphicx}
\usepackage{amsmath}
\usepackage{rotating}
\usepackage{adjustbox}
\usepackage{lscape}
\usepackage{subfig}
\usepackage{xcolor}
\usepackage{comment}
\usepackage{algorithm}
\usepackage[noend]{algpseudocode}
\usepackage{amssymb}
\usepackage{afterpage}
\usepackage[hyphens]{url}

\begin{document}
\title{Random Walk-steered Majority Undersampling\thanks{This work has emanated from research supported in part by TCG Crest, Kolkata, India and a grant from Science Foundation Ireland under Grant number [16/RC/3835]. For the purpose of Open Access, the author has applied a CC BY public copyright licence to any Author Accepted Manuscript version arising from this submission.}}
\titlerunning{RWMaU}
\author{Payel Sadhukhan\inst{1}\orcidID{0000-0001-7795-3385} \and Arjun Pakrashi \inst{2}\orcidID{0000-0002-9605-6839} \and Brian Mac\ Namee \inst{2}\orcidID{0000-0003-2518-0274}}
\authorrunning{P. Sadhukhan et al.} 
%
\institute{TCG Crest Kolkata, India \\ \email{payel0410@gmail.com}\and School of Computer Science, University College Dublin, Ireland \email{\{arjun.pakrashi,brian.macnamee\}@ucd.ie} }
\maketitle              
\begin{abstract}
In this work, we propose \emph{Random Walk-steered Majority Undersampling} (RWMaU), which undersamples the majority points of a class imbalanced dataset, in order to balance the classes. Rather than marking the majority points which belong to the neighborhood of a few minority points, we are interested to perceive the closeness of the majority points to the minority class. Random walk, a powerful tool for perceiving the proximities of connected points in a graph, is used to identify the majority points which lie close to the minority class of a class-imbalanced dataset. The visit frequencies and the order of visits of the majority points in the walks enable us to perceive an overall closeness of the majority points to the minority class. The ones lying close to the minority class are subsequently undersampled. Empirical evaluation on 21 datasets and 3 classifiers demonstrate substantial improvement in performance of RWMaU over the competing methods.  
\end{abstract}

\section{Introduction}\label{sec:introduction}
Real-world data from the domain of medical \cite{medical2}, text \cite{text1}, software defect prediction \cite{defect}, and fraud detection \cite{cyber} often have significant imbalance between target classes. In a binary classification dataset with class-imbalance, the class with more instances and the class with fewer instances are known as the majority and the minority class respectively. When a classifier is modelled on an imbalanced dataset, it often gets influenced to predict the majority class.

There are a number of different solutions to the class-imbalance problem in the literature. These can be categorized into: i) algorithmic methods \cite{method-survey}, ii) data preprocessing \cite{data-survey} and iii) ensemble-based learning \cite{ensem-survey}. Data preprocessing is the most popular choice amongst these three as it is independent of model building. In data pre-processing, we \emph{undersample} (remove points from) the majority class or \emph{oversample} (add points to) the minority class in order to reduce the difference in representation between the two classes. Consequently, the bias towards the majority class is reduced. 

The state-of-the-art undersampling schema primarily mark the majority points to be undersampled in one of two ways: i) finding a good representative set of the majority class by employing procedures like clustering, thereby discarding the majority points which lie near the clusters' periphery, or ii) marking the majority points which lie in the k-nearest neighborhood of the minority points. The k-nearest neighborhood based undersampling methods are generally focused to find the majority points which lie close to the minority points. While doing so, the methods somewhat become oblivious to the points' relative distances. Moreover, employing a crisp neighborhood-based protocol delivers a locally optimized nearness. In most cases, the methods are not motivated to quantify the overall closeness of the majority points to the minority class. 

In this work, we present \textbf{R}andom \textbf{W}alk-steered \textbf{Ma}jority \textbf{U}ndersampling (RWMaU), an undersampling technique which addresses these concerns. Instead of simply figuring out the k-nearest majority neighbors of the minority points, we are motivated to obtain the overall closeness of the majority points to the minority class. We employ random walk for this purpose. Random walk is a powerful tool for perceiving the mutual proximity of the nodes in a graph. It has been extensively used in the domain of social network analysis to find communities, compute feature representations, and find other relevant parameters of a graph \cite{random1,random2}. RWMaU forms a directed graph from a class-imbalanced dataset, where each point is connected to its k-nearest neighbors. The edge-weights of the outward edges depend on the relative distances of the neighbors. We simulate a number of random walks from the minority points (as starting nodes) and study the visit frequencies of the majority nodes (along with the order of the visits) in these walks.

In particular, we use the visits and their orders to compute the proximity scores of the majority points with respect to the minority class, thereby finding the majority points which are close to the minority class as a whole. While undersampling the majority class, we remove the majority points in order of their decreasing proximity scores. A majority point which has the highest proximity to the minority class is removed first. 

The key aspects of our work are summarized as follows.
\begin{itemize}
    \item We quantify the majority nodes' visit frequencies and the order of the visits in random walks to compute the nearness of the majority points to the minority class.
    \item The majority points which lie close to the minority class are removed to alter the class distributions in favor of the under-represented minority class.
   \item An empirical study involving 21 datasets, 3 classifiers, 5 competing methods (4 undersampling methods and the original datasets) and 2 evaluating metrics indicates the effectiveness of the proposed method. 
\end{itemize}
The remainder of the paper is organized as follows. In Section \ref{sec:related_work}, we discuss the relevant existing work in the field of handling class-imbalance. We present the random walk preliminaries, motivation of our work, and the proposed algorithm in Section \ref{sec:proposedmethod}. The experimental design is described in Section \ref{sec:experiment} and the results of the experiments are discussed in Section \ref{sec:results}. Finally, in Section \ref{sec:conclusion} we conclude the paper.

\section{Related Work}\label{sec:related_work}
One of the early approaches in the field of class-imbalance learning is algorithm-based methods. Most of the schema from this domain are motivated to either shift the class boundary away from the minority class \cite{boundary-2014} or to add a cost-sensitive learning framework where the misclassification cost of the minority class is increased \cite{cost-survey,khan-cost,wong-cost}. 
Other important classes of algorithm-based methods are active learning \cite{active-2018}, multi-objective optimization based methods \cite{MO-2020}, kernel-based methods \cite{kernel-2015} and one class classifiers \cite{oneclass-2018}.

Data-preprocessing techniques form an important and popular choice to address class-imbalance of data. In undersampling, the points belonging to the majority class are selected and removed to reduce the difference in cardinalities of the two classes. Various techniques are proposed by the researchers in this domain to make a judicious choice of the majority points to be removed from the dataset. The two principal techniques to choose points to be undersampled are i) cluster based - clustering is done to recognize the key points to be kept for the classification phase \cite{under-clus1,under-clus2,under-clus3} and ii) nearest neighbor based - the majority neighbors of the minority points are identified and are removed (with some additional condition checks) \cite{tomek,tomek1}. Oversampling of the minority class is another way of balancing the cardinalities of the two classes \cite{smote,smote-survey}. A number of diversified parameters like minority class density \cite{adasyn}, oversampling near boundary \cite{mwmote}, majority class non-encroachment \cite{sadhukhan} are considered by the researchers to effectively oversample synthetic minority points in the feature space. In recent years, random walk is learnt on graphs to choose the locations of minority oversampling \cite{rwo-oversample}. In addition to these, hybrid data-preprocessing techniques also exit in literature which employs both minority oversampling and majority undersampling \cite{hybrid1}. In some techniques, both data pre-processing and algorithm adaptation are considered to tackle the issue of class-imbalance.

The third category of class-imbalance learner deals with a set of classifiers (often at various hierarchies) along with boosting and bagging to obtain an improved learning over class-imbalanced datasets \cite{ensem-survey}. Minority oversampling is integrated with boosting to obtain an improved accuracy for both the minority and the majority class by \cite{guo-victor}. In \cite{peng-yao}, the authors follow a hierarchical paradigm where a set of weak (preliminary) classifiers are trained on the imbalanced dataset followed by derivation of a strong classifier from these weak classifiers.

\section{Random Walk-steered Majority Undersampling (RWMaU)}\label{sec:proposedmethod}

In this section, first we briefly explain related aspects of random walks followed by a brief discussion of the motivation and the core idea of our approach. Then we present the proposed algorithm, Random Walk-steered Majority Undersampling (RWMaU).

\subsection{Random Walk}

A random walk is a sequence of discrete, finite length steps in random directions depending on probabilities. Random walks are often considered in context of a graph, $G (V, E)$ where we have a set of nodes $V=\{v_{1},\dots,v_{N}\}$ and a set of edges, $E = \{(v_i,v_j)|(v_i,v_j)\in V\times V\textrm{ and } i\neq j\}$ connecting the nodes. Each edge has a weight $p_{ij}$ which connects $v_i$ and $v_j$, which can be captured in an adjacency matrix of the graph. When we consider a random walk in a graph, we start from a node, $v_{i}$, and move to another node $v_{j}$ with considering $p_{ij}$ as the transition probability. This process of moving from one note to another node is repeated until we have performed a certain number of steps. The sequence of nodes which this process goes through is called a random walk. Details about random walk can be found in \cite{lovasz}. We use properties of random walk in our proposed method.

\subsection{Motivation and Overview}

    Our approach is to mark and remove the majority points which are close to the minority class overall. To do this, we compute a score for each majority point, which determines how close they are with respect to the minority points collectively. The majority points with high scores will indicate their closeness to the minority space. This score is ranked, and the higher scored majority points are removed. Random walk serves as the backbone of this entire procedure.


We construct a directed weighted graph from the given dataset on which the random walks will be performed. We assume that the majority points, which appear a) more frequently in the walk, and, b) earlier in the walk sequence, are closer to the minority class. Based on these two assumptions the scores for each majority points are assigned. We simulate a series of random walks from each minority point and record the visit frequencies of the majority points in the series of these walks to address the assumption (a). Also, the visit frequencies of a majority node is weighted based on how far in the walk the node was visited to incorporate the assumption (b). Therefore, a visit occurring earlier in the walk will be given more weight than a later one.

A proximity score of each majority point is computed from these two information, which will indicate a degree of closeness of the majority point relative to the minority class. These assigned scores are used to identify and remove the majority points.

\subsection{Algorithm}

\begin{algorithm}[ht]
\caption{RWMaU}\label{alg:rwmau}
\begin{algorithmic}[1]
\Procedure{RWMaU}{$\mathcal{D}=\{ (\mathbf{x}_{i}, y_i), i=1,\dots,n\}$, $\alpha$, $\gamma$, $k$} 

\State $\mathcal{X}_{min} = \{\mathbf{x}_i|\forall_i y_{i}=1\}$ and $\mathcal{X}_{maj} = \{\mathbf{x}_i|\forall_i y_{i}=0\}$
\State Make graph $G(\mathcal{X}=\{\mathbf{x}_i|i=1,2,\ldots,n\}, E=\{p_{ij}|i,j=1,2,\ldots,n\})$ use Eq. (\ref{eq:pij})
\For{$\mathbf{x}_i\in\mathcal{X}_{min}$}
    \State $W_{\mathbf{x}_{i}} = randomWalk (G, \mathbf{x}_{i}, \gamma)$ (use Eq. (\ref{eq:I_eq}))
\EndFor
\For{$\mathbf{x}_l\in\mathcal{X}_{maj}$}
    \State $\nu_{l} = \sum_{\mathbf{x}_{j}\in\mathcal{X}_{min}}\sum_{\beta=1}^{\gamma}\frac{W_{\mathbf{x}_j}(\beta, \mathbf{x}_{l})}{\beta}$
\EndFor
\State $u = (|\mathcal{X}_{maj}| - |\mathcal{X}_{min}|)\times\alpha$
\State $\mathbf{\tau} = sortDecresing (\mathbf{\nu})$ \Comment{Get sorted order of $\nu$} 
\State $\mathcal{X}_{rem} = \{\mathbf{x}_{\tau_j}|\forall_{j} \nu_{\tau_j} \ge \nu_{\tau_u}\}$ \Comment{Select top $u$ points to remove}
\State $\mathcal{A} = \mathcal{X} - \mathcal{X}_{rem}$
\State \textbf{return} $\mathcal{A}$
\EndProcedure  
\end{algorithmic}
\end{algorithm}

In this section, we present Random Walk-steered Majority Undersampling (RWMaU). We will also describe the algorithm in details along with Algorithm \ref{alg:rwmau}, which summarises the scheme.

First, we represent the dataset is represented as a directed weighted graph $G (\mathcal{X}, E)$. In $G$, each vertex represents a point and the weight of the directed weighted edge from $\mathbf{x}_i$ to $\mathbf{x}_j$ is $p_{ij}$ which is defined as

\begin{equation}
    G (\mathcal{X}=\{\mathbf{x}_i|i=1,2,\ldots,n\}, E=\{p_{ij}|i,j=1,2,\ldots,n\})
\end{equation}

The $p_{ij}$ indicates the reachability of $\mathbf{x}_{j}$ from $\mathbf{x}_{i}$ in the graph and will be used as the transition probabilities in the random walk. We define $p_{ij}$ as follows







\begin{equation}\label{eq:pij}
p_{ij} =
\begin{cases}
\frac{e^{-{\frac{d_{ij}}{d_{iNN_k(i)}}}}}{\sum_{m=1}^{k}{e^{-{\frac{d_{iNN_m(i)}}{d_{iNN_k(i)}}}}}} &, \text{if } \mathbf{x}_{j} \textrm{ is a k-nearest neighbor of } \mathbf{x}_{i}   \\
\\
0 &, \text{otherwise}
\end{cases}
\end{equation}

\noindent here, $d_{ij}$ is the Euclidean distance between $\mathbf{x}_{i}$ and $\mathbf{x}_{j}$, and $d_{iNN_m(i)}$ denotes the distance between $\mathbf{x}_i$ and its $m^{th}$ nearest neighbour. 

We start a walk from each of the minority point $\mathbf{x}_{i} \in \mathcal{X}_{min}$ and record the nodes visited during the walk along with the order of visit. The random walk starting at $\mathbf{x}_{i}\in\mathcal{X}_{min}$ is represented as $W_{\mathbf{x}_{i}}$ and is defined as follows

\begin{equation}\label{eq:I_eq}
W_{\mathbf{x}_{i}}(\beta, \mathbf{x}_{l}) =
\begin{cases}
1 & \text{if } \mathbf{x}_{l} \textrm{ is visited in }\beta^{th} \textrm{ step of a walk started at } \mathbf{x}_{i} \in \mathcal{X}_{min}\\
0 & \text{otherwise}
\end{cases}
\end{equation}

\noindent here $W_{\mathbf{x}_{i}}(\beta,l)$ indicates if node $\mathbf{x}_{i}$ is visited in the $\beta^{th}$ step of a random walk started at minority instance $x_{i} \in \mathcal{X}_{min}$. The datapoints in $\mathcal{X}_{maj}$ through which relatively more random walks go through, are the ones which are more likely to be removed during the undersampling process. 
Now we can calculate the minority proximity scores of the majority instances $\mathbf{x}_{l}\in\mathcal{X}_{maj}$. We denote the minority proximity score of $\mathbf{x}_{l}$ by $\nu_{l}$, which integrates the information about visit frequency and order of $\mathbf{x}_{l}$ in the different random walks.

\begin{equation}\label{eq:nu}
\nu_{l} = \sum_{\mathbf{x}_{j}\in\mathcal{X}_{min}}\sum_{\beta=1}^{\gamma}\frac{W_{\mathbf{x}_j}(\beta, \mathbf{x}_{l})}{\beta}
\end{equation}

Once we have computed the values of $\nu_{l}$ for all majority datapoints, we can then remove the ones which with a high value of $\nu_{l}$.

We will sort the elements of $\mathcal{X}_{maj}$ in decreasing order of their $\nu$ values. We will discard the first $u$ points to get the set to remove $\mathcal{X}_{rem}$

\begin{equation}\label{eq:u}
    u=(|\mathcal{X}_{maj}|-|\mathcal{X}_{min}|)\times \alpha
\end{equation}

In Eq. (\ref{eq:u}), let $u$ be the number of points to be undersampled and $\alpha$ be a constant such that $0 < \alpha \leq 1$. $\alpha=0$ signifies no undersampling, and if we set $\alpha=1$, we will equate the cardinalities of the minority class and the majority class in the augmented set. We further denote the removed points from the majority class by $\mathcal{X}_{rem}$. Finally, the augmented training $\mathcal{A}$ set is obtained by removing the set of points to be removed through $A=\mathcal{X}-\mathcal{X}_{rem}$. $A$ is used to train the classifier.


\section{Experimental Design}\label{sec:experiment}



To evaluate the proposed method RWMaU, we have performed a detailed experiment involving 21 binary classification datasets with different degrees of class imbalance (Imbalance ratio ranging from 1.54 to 32.73). They are listed in Table \ref{tab:datasets} along with their basis statistics, where \emph{n} is the number of datapoints, \emph{d} is the number of dimensions and \emph{Imb. Ratio} is the imbalance ratio of the dataset, which is the ratio of the number of majority class and minority class datapoints. The datasets are a part of \cite{FERNANDEZ20082378,FERNANDEZ2009561} obtained from the KEEL project page \footnote{https://sci2s.ugr.es/keel/imbalanced.php}.

In the comparative study, we have included the original training dataset (without any oversampling or undersampling). Since majority class undersampling is the essence of the proposed work, we have included four undersampling schemes in this study namely -- Random Undersampling (RUS) \cite{review2}, Instance Hardness Threshold (IHT) \cite{smith2014instance}, Undersampling with Cluster Centroids (CC) and Neighbourhood Cleaning Rule (NCR) \cite{tomek1}.
K-Nearest Neighbour (with k=5), C4.5 and C4.5 + Bagging classifiers were used to train the model at their default settings were used to train the model using various undersampling schemes. The original (unsampled) dataset's performance on the above given classifier were also compared as a baseline. The value of $k$ (in RWMaU) was selected from a range of ${2,3,\ldots,10}$ and $\gamma$ (in RWMaU) was selected from a range of $2k\pm 3$ in combination by optimizing over C4.5 Decision Tree classifier. The ($k$, $\gamma$) tuple which optimized the results of RWMaU on C4.5 Decision Tree were used in all the experiments.
Also, the value of  $\alpha$ was set to $0.5$ for all the undersampling methods in the comparative study. This was done to limit removal of too many majority points.

\begin{table}[ht]
\centering
\tiny
\caption{Description of datasets}\label{tab:datasets}
\resizebox{0.4\textwidth}{!}{%
\begin{tabular}{lrrr}
  \hline
                &    n & d  & Imb. Ratio \\ 
  \hline                               
  yeast5        & 1484 & 8  & 32.73     \\ 
  yeast1289v7   & 947  & 8  & 30.57     \\ 
  wine-red-4    & 1599 & 11 & 29.17     \\ 
  yeast4        & 1484 & 8  & 28.10     \\ 
  yeast1458v7   & 693  & 8  & 22.10     \\ 
  abalone9-18   & 731  & 8  & 16.40     \\ 
  ecoli4        & 336  & 7  & 15.80     \\ 
  led02456789v1 & 443  & 7  & 10.97     \\ 
  page-blocks0  & 5472 & 10 & 8.79      \\ 
  ecoli3        & 336  & 7  & 8.60      \\ 
  yeast3        & 1484 & 8  & 8.10      \\ 
  new-thyroid1  & 215  & 5  & 5.14      \\ 
  new-thyroid2  & 215  & 5  & 5.14      \\ 
  vehicle3      & 846  & 18 & 2.99      \\ 
  vehicle1      & 846  & 18 & 2.90      \\ 
  vehicle2      & 846  & 18 & 2.88      \\ 
  glass0        & 214  & 9  & 2.06      \\ 
  pima          & 768  & 8  & 1.87      \\ 
  glass1        & 214  & 9  & 1.82      \\ 
  wdbc          & 569  & 30 & 1.68      \\ 
  spam          & 4597 & 57 & 1.54      \\ 
   \hline
\end{tabular}
}
\end{table}

For each dataset, 80\% of the points were selected randomly for training and the remaining 20\% is used for testing. The training set was used with the sampling algorithms to get the undersampled dataset. This undersampled dataset was used to train the models using the previously mentioned algorithms. We have also used the original training dataset in the empirical study and reported its outcomes. The remaining 20\% test datapoints were used to compute the model performance. The above process was repeated $10$ times and the average AUC and F1-Scores were reported and compared. The same training-test partitions were used for all the competing methods and run on the same platform .

\section{Results}\label{sec:results}

The results of the experiments are shown in Table \ref{tab:auc_tab} and \ref{tab:f1_tab} respectively. The values in the table are the mean AUC (Table \ref{tab:auc_tab}) and mean F1-Score (Table \ref{tab:f1_tab}) of the ten runs of the experiment, as mentioned in Section \ref{sec:experiment}. The values in the parentheses are the relative ranking for a sampling method and algorithm combination on the specific dataset. For example, for \emph{ecoli4} dataset, when the proposed algorithm is used with kNN, attained an AUC value of $0.9963$ and an rank of $1$, when compared with kNN used with other sampling methods and original dataset. The last row of each table shows the average rank over all datasets for a specific sampling algorithm and classification algorithm pair.

The objective is to  the relative efficacy of RWMaU in learning imbalanced datasets as compared to the competing paradigms. The comparison is done for each sampling method and classifier pair. With respect to kNN classifier, on both metrics, RWMaU did very well compared to other undersampling methods as well as the original dataset. RWMaU has also performed best on C4.5 Decision Tree with an average rank of 1.48. Particularly, the difference of average ranks of RWMaU and the next based ranked method is remarkable. For (Bagging + C4.5), RWMaU achieved the lowest average rank on both minority class F1 and AUC scores. However, it is worth noting that the difference with respect to next best average rank was not that significant as for the previous two cases. In case of minority F1 score, the thresholding was done using 0.5, we find RWMaU performing better overall with respect to the different classifiers. 





\begin{figure}[ht]
\centering
\subfloat[AUC C4.5\label{fig:auc_c45}]{\includegraphics[width=.45\textwidth]{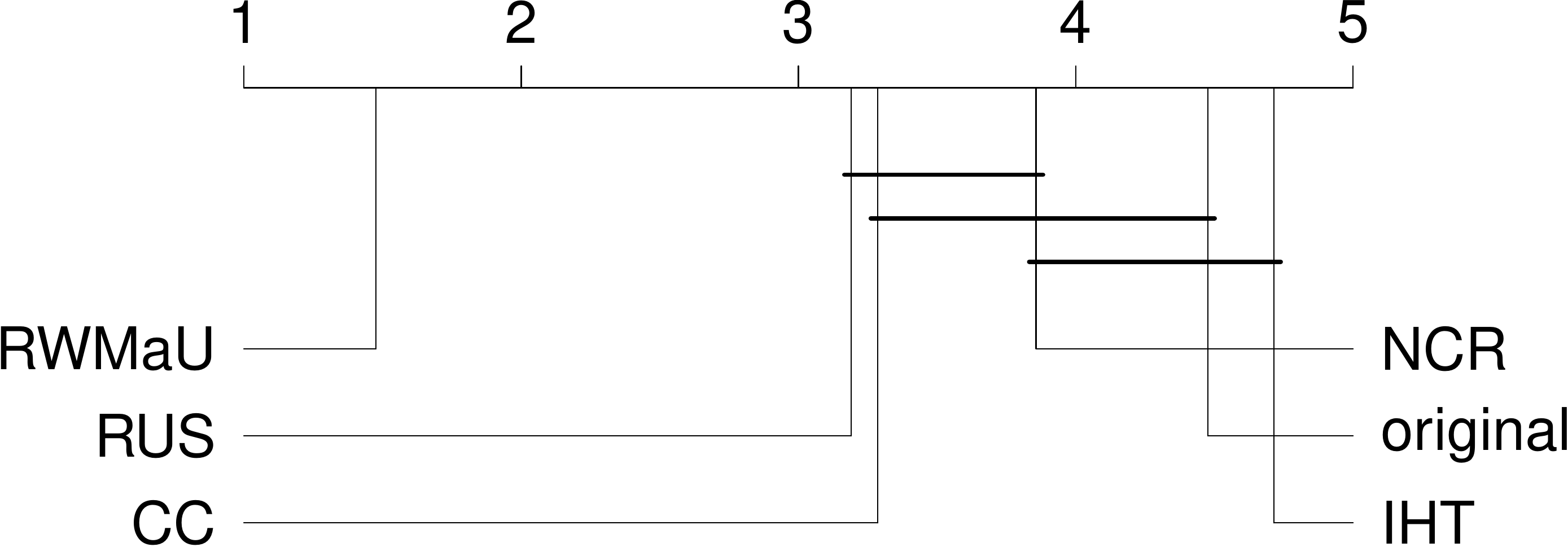}}\ \ \ \ \subfloat[F1 Score C4.5\label{fig:f1_c45}]{\includegraphics[width=.45\textwidth]{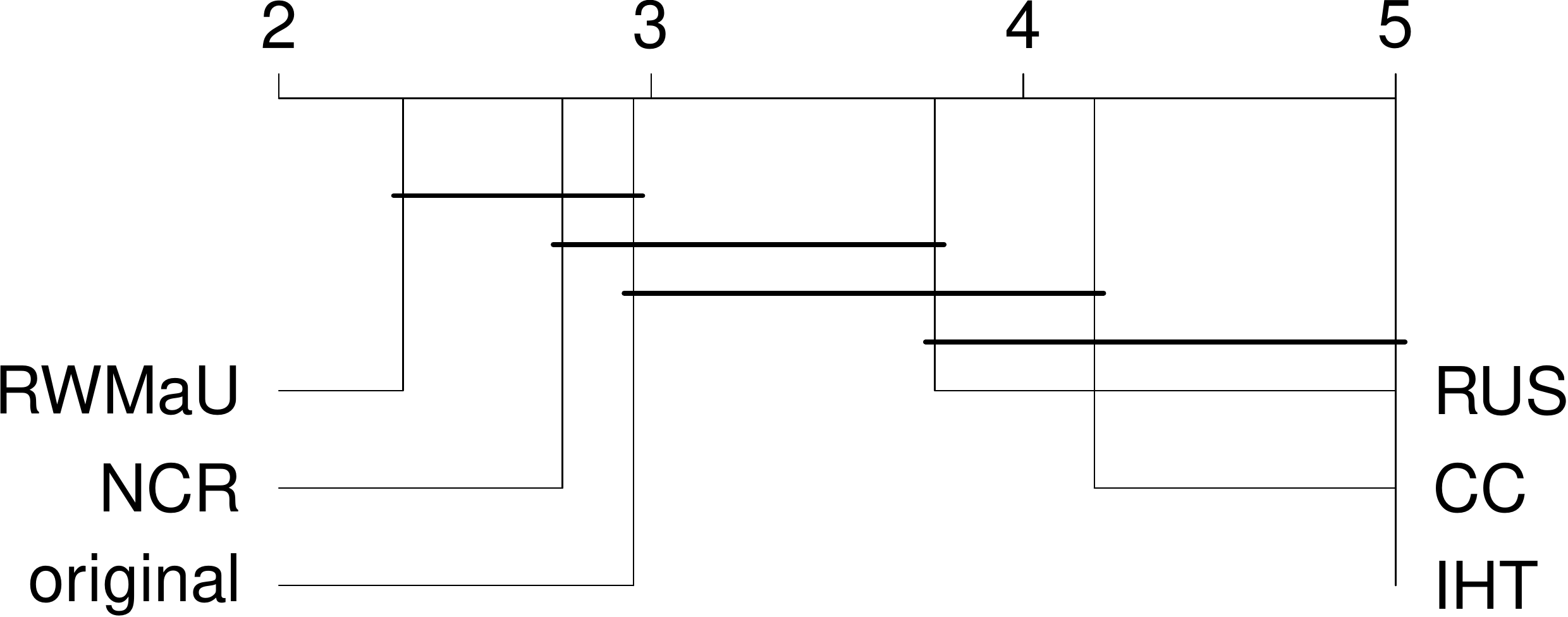}}

\subfloat[AUC C4.5 + Bagging\label{fig:auc_c45_bag}]{\includegraphics[width=.45\textwidth]{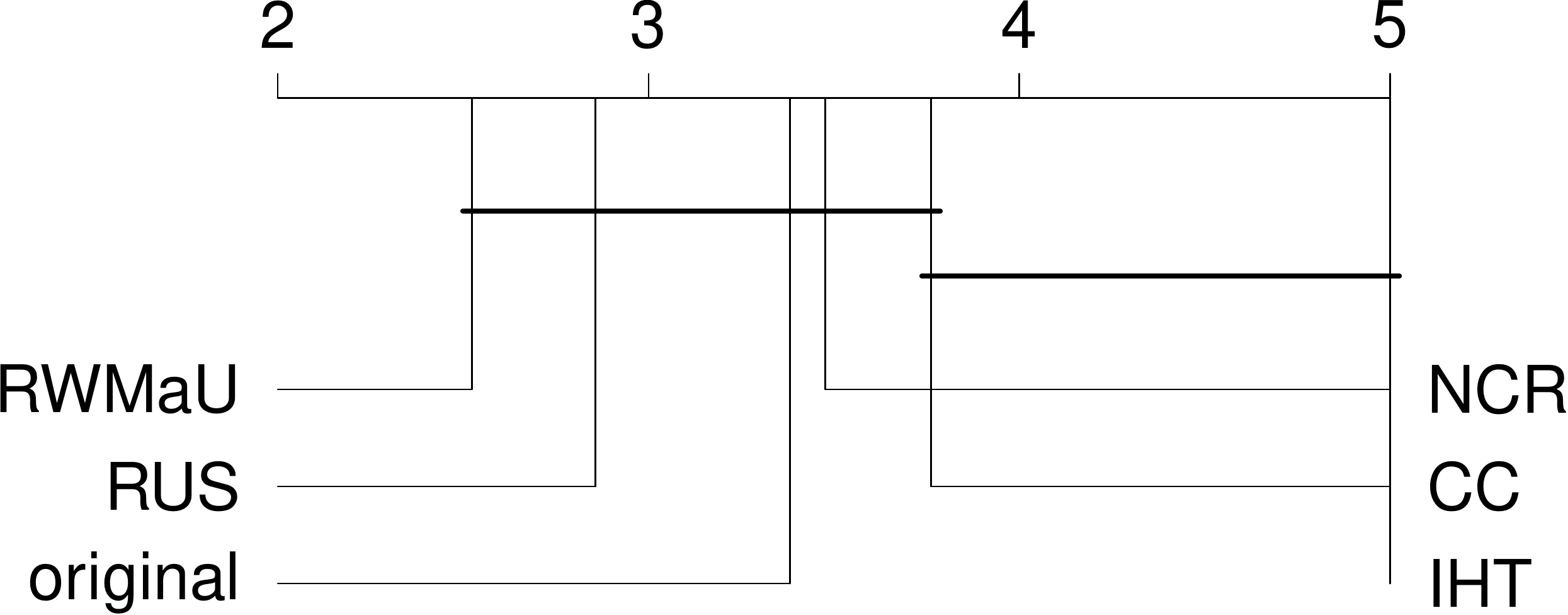}}\ \ \ \ \subfloat[F1 Score C4.5 + Bagging\label{fig:f1_c45_bag}]{\includegraphics[width=.45\textwidth]{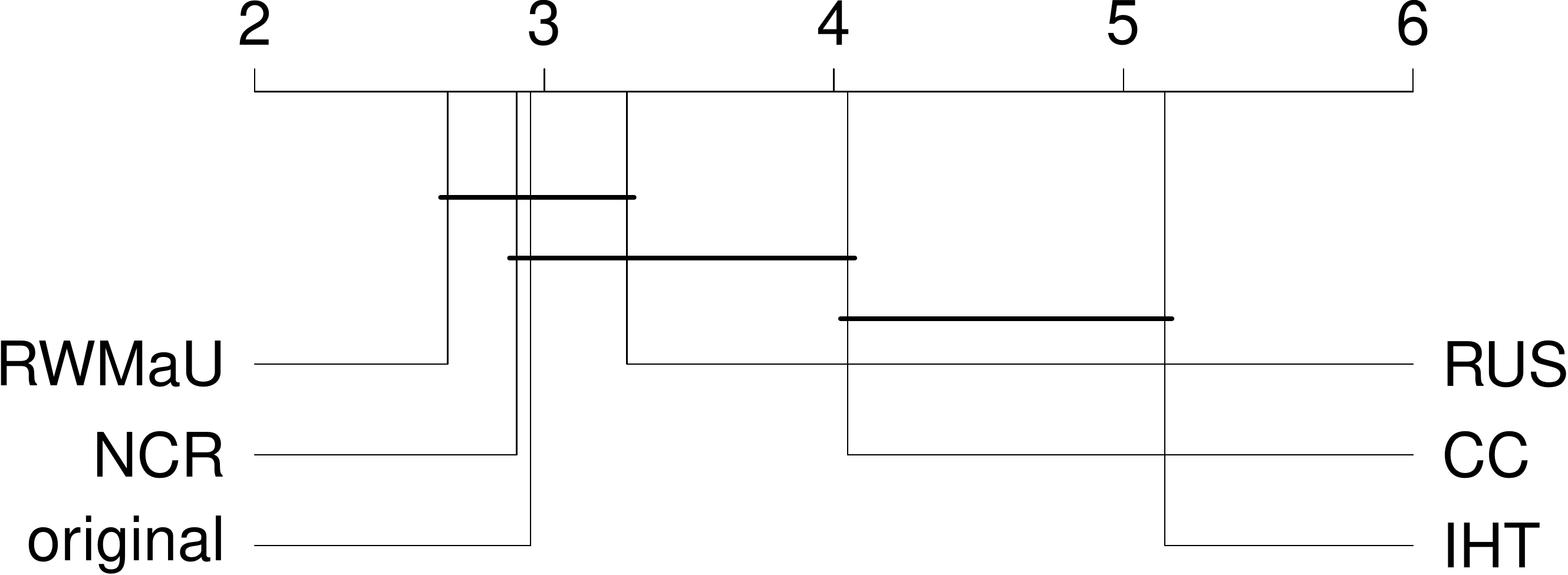}}

\subfloat[AUC KNN\label{fig:auc_knn}]{\includegraphics[width=.45\textwidth]{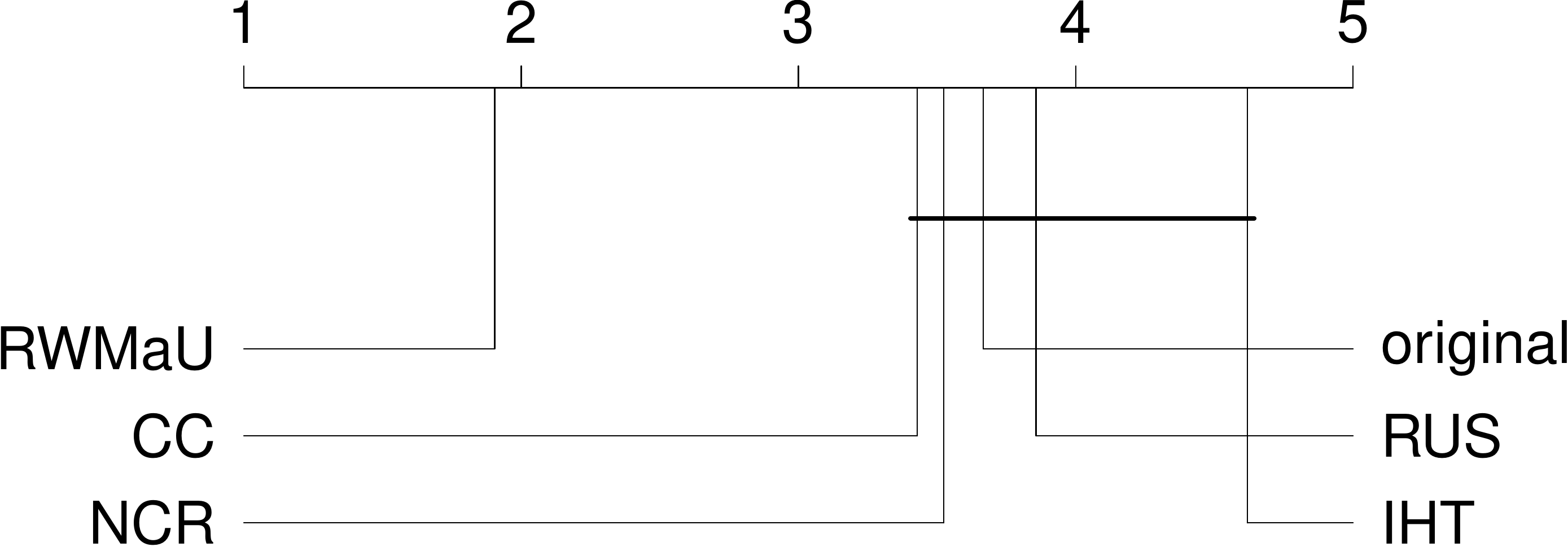}}\ \ \ \ \subfloat[F1 Score KNN\label{fig:f1_knn}]{\includegraphics[width=.45\textwidth]{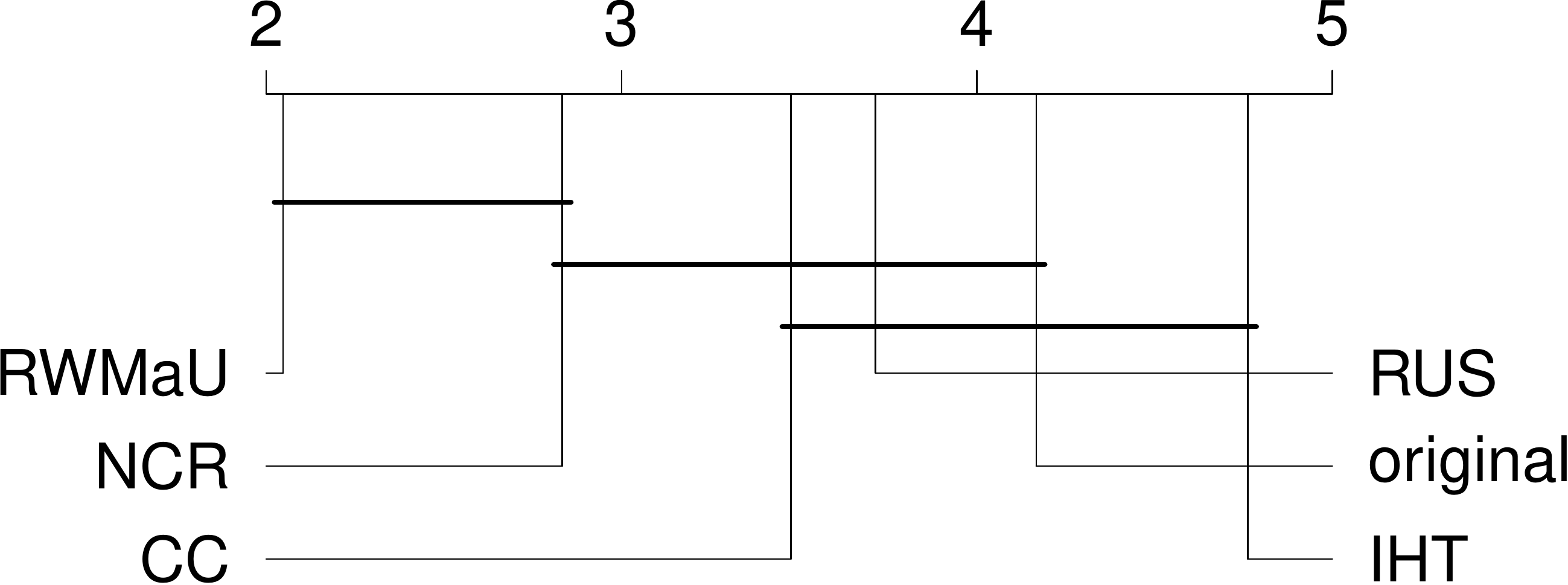}}

\caption{Critical difference plots of the post-hoc Friedman test with significance level $\alpha = 0.05$ for AUC and F1 Score from the KNN, C4.5 and C4.5 + Bagging experiments.\label{fig:cdplots}}

\end{figure}

To further investigate, we have performed a statistical significance tests to understand the pairwise differences between the methods. The Friedman test with Finner $p$-value correction was performed to do a multiple classifier test on each combination of sampling method, used algorithm, on both F1-Score and AUC scores for each sampling and classifier combination. The summary results are represented as critical difference plots in Figure \ref{fig:cdplots} \footnote{The full result tables in supplementary material: \url{https://github.com/phoxis/rwmau/blob/main/RWMaU_ICONIP2021_Paper_Supplementary_Material.pdf}.}.

With respect to AUC scores, from the critical difference plots, it is clear that RWMaU performed significantly better than all the methods with used with C4.5 in Figure \ref{fig:auc_c45} and KNN in Figure \ref{fig:auc_knn}, with a significance level of $\alpha = 0.05$. Although the null hypothesis could not be rejected, except for IHT for C4.5 + Bagging case as can be seen in Figure \ref{fig:auc_c45_bag}.

When we consider the same for F1 Score, RWMaU's performance was found to be significantly better than RUS, CC and IHT with C4.5 (Figure \ref{fig:f1_c45}) and all other method except NCR with KNN (Figure \ref{fig:f1_knn}). On the other hand, for C4.5 + Bagging (Figure \ref{fig:f1_c45_bag}), RWMaU was found to be significantly better than RUS, CC and IHT. In general, we may conclude that, RWMaU would improve the learning of the class-imbalanced datasets over the competing methods.

\section{Conclusion}\label{sec:conclusion}
In this paper, we address the class imbalance problem by proposing an undersampling method, Random Walk-steered Majority Undersampling (RWMaU). Our scheme re-balances the dataset by removing datapoints from the majority class. The main objective is to remove the majority points which are relatively closer to the minority class. 
The novelty of our method lies in the use of random walk visits to perceive the nearness of the points in a dataset. The majority class points which lie close to the minority class are subsequently undersampled.
The AUC scores and the minority class F1 scores obtained from our empirical study show that RWMaU delivers improved performance over existing methods. RWMaU + KNN and RWMaU + C4.5 performed significantly better than all the other methods with respect to AUC scores, whereas they performed significantly better than most of the methods with respect to F1 score. Overall, RWMaU has attained the best rank in all cases with respect to both AUC and F1 scores. In future, we would like to design a minority-oversampling scheme which is built upon the random walks over the instances of a class-imbalanced dataset. It would also be interesting to integrate random walk based oversampling and undersampling in a single framework.







\bibliographystyle{splncs04}
\bibliography{rwalksamp}

\begin{thebibliography}{10}
\providecommand{\url}[1]{\texttt{#1}}
\providecommand{\urlprefix}{URL }
\providecommand{\doi}[1]{https://doi.org/#1}

\bibitem{mwmote}
Barua, S., Islam, M.M., Yao, X., Murase, K.: {MWMOTE}--majority weighted
  minority oversampling technique for imbalanced data set learning. IEEE
  Transactions on Knowledge and Data Engineering  \textbf{26}(2),  405--425
  (Feb 2014)

\bibitem{defect}
Chakraborty, T., Chakraborty, A.K.: Hellinger net: A hybrid imbalance learning
  model to improve software defect prediction. IEEE Transactions on Reliability
   \textbf{70}(2),  481--494 (2021)

\bibitem{smote}
Chawla, N.V., Bowyer, K.W., Hall, L.O., Kegelmeyer, W.P.: Smote: Synthetic
  minority over-sampling technique. J. Artif. Int. Res.  \textbf{16}(1),
  321--357 (Jun 2002)

\bibitem{tomek}
Devi, D., Purkayastha, B., et~al.: Redundancy-driven modified tomek-link based
  undersampling: A solution to class imbalance. Pattern Recognition Letters
  \textbf{93},  3--12 (2017)

\bibitem{smote-survey}
Elreedy, D., Atiya, A.F.: A comprehensive analysis of synthetic minority
  oversampling technique {(SMOTE)} for handling class imbalance. Information
  Sciences  \textbf{505},  32--64 (2019)

\bibitem{FERNANDEZ2009561}
Fernández, A., {del Jesus}, M.J., Herrera, F.: Hierarchical fuzzy rule based
  classification systems with genetic rule selection for imbalanced data-sets.
  International Journal of Approximate Reasoning  \textbf{50}(3),  561--577
  (2009), special Section on Bayesian Modelling

\bibitem{FERNANDEZ20082378}
Fernández, A., García, S., {del Jesus}, M.J., Herrera, F.: A study of the
  behaviour of linguistic fuzzy rule based classification systems in the
  framework of imbalanced data-sets. Fuzzy Sets and Systems  \textbf{159}(18),
  2378--2398 (2008), theme: Information Processing

\bibitem{ensem-survey}
Galar, M., Fernández, A., Barrenechea, E., Bustince, H., Herrera, F.: A review
  on ensembles for the class imbalance problem: bagging-, boosting-, and
  hybrid-based approaches. IEEE Transactions on Systems, Man, and Cybernetics,
  Part C (Applications and Reviews)  \textbf{42}(4),  463--484 (2011)

\bibitem{guo-victor}
Guo, H., Viktor, H.L.: Learning from imbalanced data sets with boosting and
  data generation: the databoost-im approach. ACM Sigkdd Explorations
  Newsletter  \textbf{6}(1),  30--39 (2004)

\bibitem{data-survey}
Haixiang, G., Yijing, L., Shang, J., Mingyun, G., Yuanyue, H., Bing, G.:
  Learning from class-imbalanced data: Review of methods and applications.
  Expert Systems with Applications  \textbf{73},  220--239 (2017)

\bibitem{adasyn}
He, H., Bai, Y., Garcia, E.A., Li, S.: {ADASYN}: Adaptive synthetic sampling
  approach for imbalanced learning. In: 2008 IEEE international joint
  conference on neural networks (IEEE world congress on computational
  intelligence). pp. 1322--1328. IEEE (2008)

\bibitem{random1}
Jamali, M., Ester, M.: Trustwalker: a random walk model for combining
  trust-based and item-based recommendation. In: Proceedings of the 15th ACM
  SIGKDD international conference on Knowledge discovery and data mining. pp.
  397--406 (2009)

\bibitem{random2}
Katzir, L., Hardiman, S.J.: Estimating clustering coefficients and size of
  social networks via random walk. ACM Transactions on the Web (TWEB)
  \textbf{9}(4),  1--20 (2015)

\bibitem{method-survey}
Kaur, H., Pannu, H.S., Malhi, A.K.: A systematic review on imbalanced data
  challenges in machine learning: Applications and solutions. ACM Computing
  Surveys (CSUR)  \textbf{52}(4),  1--36 (2019)

\bibitem{khan-cost}
Khan, S.H., Hayat, M., Bennamoun, M., Sohel, F.A., Togneri, R.: Cost-sensitive
  learning of deep feature representations from imbalanced data. IEEE
  transactions on neural networks and learning systems  \textbf{29}(8),
  3573--3587 (2017)

\bibitem{oneclass-2018}
Krawczyk, B., Galar, M., Wo{\'z}niak, M., Bustince, H., Herrera, F.: Dynamic
  ensemble selection for multi-class classification with one-class classifiers.
  Pattern Recognition  \textbf{83},  34--51 (2018)

\bibitem{tomek1}
Laurikkala, J.: Improving identification of difficult small classes by
  balancing class distribution. In: Conference on Artificial Intelligence in
  Medicine in Europe. pp. 63--66. Springer (2001)

\bibitem{under-clus2}
Lin, W.C., Tsai, C.F., Hu, Y.H., Jhang, J.S.: Clustering-based undersampling in
  class-imbalanced data. Information Sciences  \textbf{409},  17--26 (2017)

\bibitem{cost-survey}
Ling, C.X., Sheng, V.S.: Cost-sensitive learning and the class imbalance
  problem. Encyclopedia of machine learning  \textbf{2011},  231--235 (2008)

\bibitem{lovasz}
Lov{\'a}sz, L.: Random walks on graphs. Combinatorics, Paul erdos is eighty
  \textbf{2}(1-46), ~4 (1993)

\bibitem{boundary-2014}
Maratea, A., Petrosino, A., Manzo, M.: Adjusted f-measure and kernel scaling
  for imbalanced data learning. Information Sciences  \textbf{257},  331--341
  (2014)

\bibitem{medical2}
Mena, L.J., Gonzalez, J.A.: Machine learning for imbalanced datasets:
  Application in medical diagnostic. In: Flairs Conference. pp. 574--579 (2006)

\bibitem{review2}
Mohammed, R., Rawashdeh, J., Abdullah, M.: Machine learning with oversampling
  and undersampling techniques: overview study and experimental results. In:
  2020 11th International Conference on Information and Communication Systems
  (ICICS). pp. 243--248. IEEE (2020)

\bibitem{under-clus1}
Ofek, N., Rokach, L., Stern, R., Shabtai, A.: Fast-cbus: A fast
  clustering-based undersampling method for addressing the class imbalance
  problem. Neurocomputing  \textbf{243},  88--102 (2017)

\bibitem{peng-yao}
Peng, Y., Yao, J.: {AdaOUBoost}: adaptive over-sampling and under-sampling to
  boost the concept learning in large scale imbalanced data sets. In:
  Proceedings of the international conference on Multimedia information
  retrieval. pp. 111--118 (2010)

\bibitem{hybrid1}
Ramentol, E., Caballero, Y., Bello, R., Herrera, F.: {SMOTE-RS B*}: a hybrid
  preprocessing approach based on oversampling and undersampling for high
  imbalanced data-sets using smote and rough sets theory. Knowledge and
  information systems  \textbf{33}(2),  245--265 (2012)

\bibitem{MO-2020}
Ribeiro, V.H.A., Reynoso-Meza, G.: Ensemble learning by means of a
  multi-objective optimization design approach for dealing with imbalanced data
  sets. Expert Systems with Applications  \textbf{147},  113232 (2020)

\bibitem{sadhukhan}
Sadhukhan, P.: Learning minority class prior to minority oversampling. In: 2019
  International Joint Conference on Neural Networks (IJCNN). pp.~1--8 (2019)

\bibitem{smith2014instance}
Smith, M.R., Martinez, T., Giraud-Carrier, C.: An instance level analysis of
  data complexity. Machine learning  \textbf{95}(2),  225--256 (2014)

\bibitem{kernel-2015}
Tang, B., He, H.: Kerneladasyn: Kernel based adaptive synthetic data generation
  for imbalanced learning. In: 2015 IEEE Congress on Evolutionary Computation
  (CEC). pp. 664--671. IEEE (2015)

\bibitem{cyber}
Wheelus, C., Bou-Harb, E., Zhu, X.: Tackling class imbalance in cyber security
  datasets. In: 2018 IEEE International Conference on Information Reuse and
  Integration (IRI). pp. 229--232. IEEE (2018)

\bibitem{wong-cost}
Wong, M.L., Seng, K., Wong, P.K.: Cost-sensitive ensemble of stacked denoising
  autoencoders for class imbalance problems in business domain. Expert Systems
  with Applications  \textbf{141},  112918 (2020)

\bibitem{active-2018}
Zhang, C., Tavanapong, W., Kijkul, G., Wong, J., De~Groen, P.C., Oh, J.:
  Similarity-based active learning for image classification under class
  imbalance. In: 2018 IEEE International Conference on Data Mining (ICDM). pp.
  1422--1427. IEEE (2018)

\bibitem{rwo-oversample}
Zhang, H., Li, M.: Rwo-sampling: A random walk over-sampling approach to
  imbalanced data classification. Information Fusion  \textbf{20},  99--116
  (2014)

\bibitem{under-clus3}
Zhang, Y.P., Zhang, L.N., Wang, Y.C.: Cluster-based majority under-sampling
  approaches for class imbalance learning. In: 2010 2nd IEEE International
  Conference on Information and Financial Engineering. pp. 400--404. IEEE
  (2010)

\bibitem{text1}
Zheng, Z., Wu, X., Srihari, R.: Feature selection for text categorization on
  imbalanced data. SIGKDD Explor. Newsl.  \textbf{6}(1),  80--89 (Jun 2004)

\end{thebibliography}

\begin{landscape}
\begin{table}
\centering
\caption{AUC Scores}\label{tab:auc_tab}
\resizebox{!}{0.22\textheight}{
\begin{tabular}{l|llllll|llllll|llllll}
\hline 
 & \multicolumn{6}{c|}{KNN} & \multicolumn{6}{c|}{C4.5} & \multicolumn{6}{c}{C4.5 + Bagging}\tabularnewline
\hline 
              & RWMaU   & Original   & RUS        & IHT        & CC         & NCR        & RWMaU   & Original   & RUS        & IHT        & CC         & NCR        & RWMaU & Original & RUS & IHT & CC & NCR\tabularnewline
\hline                                                        
yeast5        & 0.9871 (2) & 0.9830 (4) & 0.9735 (5) & 0.8497 (6) & 0.9868 (3) & \textbf{0.9909 (1)} & 0.9489 (3) & 0.8422 (5) & 0.9721 (2) & 0.8382 (6) & \textbf{0.9767 (1)} & 0.8647 (4) & 0.9837 (2) & 0.9548 (4) & 0.9828 (3) & 0.9241 (6) & \textbf{0.9876 (1)} & 0.9479 (5) \tabularnewline
yeast1289v7   & 0.7214 (2) & 0.6971 (3) & 0.6480 (5) & 0.5867 (6) & \textbf{0.7510 (1)} & 0.6804 (4) & \textbf{0.6051 (1)} & 0.5653 (4) & 0.5441 (6) & 0.5845 (3) & 0.5889 (2) & 0.5465 (5) & 0.6894 (3) & 0.6896 (2) & \textbf{0.7001} (1) & 0.5799 (6) & 0.6300 (5) & 0.6598 (4) \tabularnewline
wine-red-4    & 0.5871 (3) & 0.4987 (6) & 0.5765 (4) & 0.6048 (2) & \textbf{0.6321 (1)} & 0.5202 (5) & 0.5970 (2) & 0.5500 (5) & \textbf{0.6027 (1)} & 0.5325 (6) & 0.5902 (3) & 0.5585 (4) & \textbf{0.7574 (1)} & 0.6684 (5) & 0.6966 (3) & 0.6148 (6) & 0.6816 (4) & 0.7355 (2) \tabularnewline
yeast4        & 0.8481 (3) & 0.7755 (5) & 0.8906 (2) & 0.7179 (6) & \textbf{0.9094 (1)} & 0.8226 (4) & 0.7902 (2) & 0.6505 (5) & \textbf{0.8122 (1)} & 0.5825 (6) & 0.7119 (3) & 0.6899 (4) & \textbf{0.8900 (1)} & 0.7932 (5) & 0.8805 (2) & 0.6552 (6) & 0.8195 (4) & 0.8456 (3) \tabularnewline
yeast1458v7   & \textbf{0.6934 (1)} & 0.6233 (5) & 0.6079 (6) & 0.6479 (3) & 0.6535 (2) & 0.6338 (4) & \textbf{0.5645 (1)} & 0.4994 (6) & 0.5604 (2) & 0.5352 (3) & 0.5290 (4) & 0.5260 (5) & 0.6168 (4) & 0.6719 (2) & 0.6314 (3) & 0.5689 (5) & 0.5550 (6) & \textbf{0.7128 (1)} \tabularnewline
abalone9-18   & \textbf{0.7748 (1)} & 0.7307 (3) & 0.7092 (5) & 0.6271 (6) & 0.7131 (4) & 0.7328 (2) & \textbf{0.6813 (1)} & 0.6691 (3) & 0.6541 (5) & 0.5577 (6) & 0.6667 (4) & 0.6754 (2) & \textbf{0.8239 (1)} & 0.8105 (3) & 0.7979 (4) & 0.6057 (6) & 0.7787 (5) & 0.8107 (2) \tabularnewline
ecoli4        & \textbf{0.9963 (1)} & 0.9846 (5) & 0.9899 (3) & 0.9155 (6) & 0.9962 (2) & 0.9848 (4) & 0.8440 (2) & 0.8264 (5) & 0.8391 (3) & 0.8357 (4) & \textbf{0.8577 (1)} & 0.8123 (6) & \textbf{0.9315 (1)} & 0.9055 (5) & 0.9228 (3) & 0.9239 (2) & 0.9177 (4) & 0.8535 (6) \tabularnewline
led02456789v1 & 0.7555 (5) & 0.7635 (4) & 0.8205 (2) & \textbf{0.8389 (1)} & 0.7544 (6) & 0.7683 (3) & \textbf{0.6924 (1)} & 0.5726 (6) & 0.6103 (3) & 0.6277 (2) & 0.6087 (4) & 0.5809 (5) & 0.7497 (2) & 0.7206 (4) & 0.7177 (6) & 0.7180 (5) & \textbf{0.7712 (1)} & 0.7231 (3) \tabularnewline
page-blocks0  & \textbf{0.9586 (1)} & 0.9377 (4) & 0.9473 (2) & 0.5770 (6) & 0.9132 (5) & 0.9429 (3) & 0.9343 (2) & 0.8994 (4) & \textbf{0.9407 (1)} & 0.5874 (6) & 0.8752 (5) & 0.9292 (3) & 0.9703 (4) & 0.9721 (3) & \textbf{0.9798 (1)} & 0.5933 (6) & 0.9348 (5) & 0.9761 (2) \tabularnewline
ecoli3        & \textbf{0.9363 (1)} & 0.9265 (4) & 0.9301 (2) & 0.8329 (6) & 0.9232 (5) & 0.9273 (3) & \textbf{0.8264 (1)} & 0.7664 (6) & 0.7957 (4) & 0.7807 (5) & 0.8168 (2) & 0.7970 (3) & 0.8878 (4) & 0.8825 (6) & 0.8931 (3) & 0.8984 (2) & \textbf{0.9099 (1)} & 0.8863 (5) \tabularnewline
yeast3        & \textbf{0.9586 (1)} & 0.9407 (5) & 0.9463 (3) & 0.8716 (6) & 0.9574 (2) & 0.9462 (4) & \textbf{0.8969} (1) & 0.8223 (6) & 0.8931 (3) & 0.8268 (5) & 0.8966 (2) & 0.8852 (4) & 0.9588 (2) & 0.9410 (5) & \textbf{0.9608 (1)} & 0.9021 (6) & 0.9557 (3) & 0.9522 (4) \tabularnewline
new-thyroid1  & 0.9897 (3) & 0.9927 (2) & 0.9866 (4) & 0.9109 (6) & 0.9845 (5) & \textbf{0.9966 (1)} & 0.9462 (2) & 0.8857 (4) & 0.9461 (3) & 0.8418 (6) & \textbf{0.9474 (1)} & 0.8856 (5) & 0.9854 (3) & 0.9814 (4) & \textbf{0.9904 (1)} & 0.9624 (5) & 0.9882 (2) & 0.9436 (6) \tabularnewline
new-thyroid2  & 0.9724 (5) & \textbf{0.9977 (1)} & 0.9808 (4) & 0.8940 (6) & 0.9858 (3) & 0.9919 (2) & \textbf{0.9593 (1)} & 0.9158 (4) & 0.9172 (3) & 0.8751 (6) & 0.9420 (2) & 0.8920 (5) & 0.9955 (2) & \textbf{0.9965 (1)} & 0.9886 (3) & 0.9785 (6) & 0.9877 (4) & 0.9814 (5) \tabularnewline
vehicle3      & \textbf{0.7816 (1)} & 0.7742 (4) & 0.7653 (5) & 0.7758 (3) & 0.7578 (6) & 0.7788 (2) & 0.7446 (2) & 0.6855 (6) & 0.7046 (3) & \textbf{0.7514 (1)} & 0.6882 (5) & 0.7024 (4) & 0.8110 (4) & 0.8180 (2) & 0.8043 (6) & 0.8107 (5) & 0.8143 (3) & \textbf{0.8278 (1)} \tabularnewline
vehicle1      & \textbf{0.7735 (1)} & 0.7578 (4) & 0.7458 (6) & 0.7679 (2) & 0.7469 (5) & 0.7621 (3) & \textbf{0.7673} (1) & 0.6668 (6) & 0.6998 (4) & 0.7427 (3) & 0.6777 (5) & \textbf{0.7432 (2)} & 0.8310 (2) & 0.8191 (5) & 0.8214 (4) & \textbf{0.8325} (1) & 0.8052 (6) & 0.8272 (3) \tabularnewline
vehicle2      & 0.9485 (2) & \textbf{0.9600 (1)} & 0.9307 (5) & 0.9120 (6) & 0.9468 (3) & 0.9460 (4) & \textbf{0.9413 (1)} & 0.9286 (4) & 0.9213 (5) & 0.8945 (6) & 0.9385 (2) & 0.9340 (3) & 0.9861 (4) & 0.9862 (3) & \textbf{0.9884 (1)} & 0.9494 (6) & 0.9779 (5) & 0.9868 (2) \tabularnewline
glass0        & \textbf{0.8682 (1)} & 0.8491 (3) & 0.8539 (2) & 0.7704 (6) & 0.8463 (4) & 0.8423 (5) & 0.8094 (2) & \textbf{0.8373 (1)} & 0.7931 (4) & 0.7386 (6) & 0.7392 (5) & 0.8060 (3) & 0.8795 (3) & \textbf{0.9239 (1)} & 0.8894 (2) & 0.7866 (6) & 0.8531 (5) & 0.8672 (4) \tabularnewline
pima          & 0.7632 (2) & 0.7288 (6) & 0.7604 (3) & \textbf{0.7822 (1)} & 0.7451 (5) & 0.7598 (4) & 0.6114 (2) & 0.5606 (5) & 0.5659 (4) & \textbf{0.6217 (1)} & 0.5461 (6) & 0.6036 (3) & \textbf{0.6820 (1)} & 0.6404 (5) & 0.6476 (4) & 0.6728 (2) & 0.5624 (6) & 0.6650 (3) \tabularnewline
glass1        & 0.8702 (2) & \textbf{0.8844 (1)} & 0.8417 (5) & 0.7988 (6) & 0.8607 (3) & 0.8521 (4) & \textbf{0.7365 (1)} & 0.6946 (5) & 0.7054 (2) & 0.6258 (6) & 0.6966 (4) & 0.7006 (3) & 0.8392 (2) & \textbf{0.8404 (1)} & 0.8119 (4) & 0.6814 (6) & 0.8082 (5) & 0.8139 (3) \tabularnewline
wdbc          & 0.9519 (1) & 0.9454 (5) & 0.9497 (4) & 0.9504 (2) & 0.9500 (3) & 0.9431 (6) & \textbf{0.9372 (1)} & 0.9291 (2) & 0.9239 (4) & 0.9087 (6) & 0.9189 (5) & 0.9276 (3) & 0.9803 (5) & \textbf{0.9855 (1)} & 0.9840 (2) & 0.9712 (6) & 0.9826 (3) & 0.9809 (4) \tabularnewline
spam          & \textbf{0.8630 (1)} & 0.8580 (2) & 0.8503 (4) & 0.8499 (5) & 0.8510 (3) & 0.8458 (6) & \textbf{0.9069 (1)} & 0.9011 (2) & 0.8927 (4) & 0.8672 (6) & 0.8964 (3) & 0.8885 (5) & 0.9712 (2) & 0.9692 (4) & 0.9694 (3) & 0.9539 (6) & \textbf{0.9724 (1)} & 0.9644 (5) \tabularnewline
\hline        
Avg. rank              & 1.90       & 3.67       & 3.86       & 4.62       & 3.43       & 3.52       & 1.48       & 4.48       & 3.19       & 4.71       & 3.29       & 3.86       & 2.52       & 3.38       & 2.86       & 5.00       & 3.76       & 3.48       \tabularnewline
\hline 
\end{tabular}

}
\caption{F1 Scores}\label{tab:f1_tab}
\resizebox{!}{0.22\textheight}{

\begin{tabular}{l|llllll|llllll|llllll}
\hline 
 & \multicolumn{6}{c|}{KNN} & \multicolumn{6}{c|}{C4.5} & \multicolumn{6}{c}{C4.5 + Bagging}\tabularnewline
\hline 
                & RWMaU   & Original     & RUS        & IHT        & CC         & NCR          & RWMaU   & Original   & RUS        & IHT        & CC         & NCR        & RWMaU  & Original & RUS & IHT & CC & NCR\tabularnewline
\hline 
yeast5          & 0.6200 (3) & 0.6606 (2.0) & 0.3978 (5) & 0.1420 (6) & 0.4052 (4) & \textbf{0.7540 (1.0)} & 0.4662 (5) & \textbf{0.6995 (1)} & 0.5478 (4) & 0.1726 (6) & 0.6000 (3) & 0.6470 (2) & 0.4699 (5) & 0.6760 (2) & 0.5128 (4) & 0.1972 (6) & 0.5647 (3) & \textbf{0.7272 (1)} \tabularnewline
yeast1289v7     & \textbf{0.1938 (1)} & 0.0000 (5.5) & 0.0930 (3) & 0.0760 (4) & 0.1064 (2) & 0.0000 (5.5) & \textbf{0.1953 (1)} & 0.1680 (2) & 0.0795 (6) & 0.0856 (5) & 0.0886 (4) & 0.1316 (3) & 0.1688 (3) & \textbf{0.2052 (1)} & 0.1274 (4) & 0.0786 (6) & 0.0904 (5) & 0.1821 (2) \tabularnewline
wine-red-4      & \textbf{0.1345 (1)} & 0.0000 (6.0) & 0.0898 (4) & 0.0924 (3) & 0.0952 (2) & 0.0787 (5.0) & \textbf{0.1652 (1)} & 0.1235 (3) & 0.1023 (4) & 0.0782 (6) & 0.0897 (5) & 0.1246 (2) & \textbf{0.1452 (1)} & 0.0897 (4) & 0.1318 (2) & 0.0791 (5) & 0.1004 (3) & 0.0438 (6) \tabularnewline
yeast4          & \textbf{0.4530 (1)} & 0.1616 (5.0) & 0.2824 (4) & 0.1148 (6) & 0.3137 (3) & 0.4200 (2.0) & 0.1943 (4) & 0.2751 (2) & 0.2164 (3) & 0.0771 (6) & 0.1245 (5) & \textbf{0.2847 (1)} & 0.2132 (4) & 0.2570 (2) & 0.2140 (3) & 0.0761 (6) & 0.1472 (5) & \textbf{0.3013 (1)} \tabularnewline
yeast1458v7     & \textbf{0.1433 (1)} & 0.0250 (6.0) & 0.1123 (3) & 0.1033 (4) & 0.1189 (2) & 0.0450 (5.0) & 0.0947 (2) & 0.0461 (6) &\textbf{ 0.0992 (1)} & 0.0861 (3) & 0.0793 (4) & 0.0769 (5) & 0.1045 (2) & 0.0250 (5) & \textbf{0.1193 (1)} & 0.0891 (3) & 0.0863 (4) & 0.0200 (6) \tabularnewline
abalone9-18     & \textbf{0.3203 (1)} & 0.0836 (6.0) & 0.1978 (3) & 0.1339 (5) & 0.2223 (2) & 0.1479 (4.0) & 0.2978 (3) & \textbf{0.3540 (1)} & 0.1945 (5) & 0.1307 (6) & 0.2055 (4) & 0.3460 (2) & 0.3650 (3) & \textbf{0.4101 (1)} & 0.2316 (5) & 0.1323 (6) & 0.2968 (4) & 0.3882 (2) \tabularnewline
ecoli4          & \textbf{0.8770 (1)} & 0.8571 (3.0) & 0.5377 (5) & 0.3148 (6) & 0.5533 (4) & 0.8587 (2.0) & 0.4124 (5) & \textbf{0.6760 (1)} & 0.4292 (3) & 0.3160 (6) & 0.4177 (4) & 0.6419 (2) & 0.4898 (3) & \textbf{0.6698 (1)} & 0.4739 (4) & 0.3405 (6) & 0.4362 (5) & 0.6664 (2) \tabularnewline
led02456789v1   & \textbf{0.5000 (1)} & 0.4300 (4.0) & 0.4402 (3) & 0.4072 (6) & 0.4249 (5) & 0.4467 (2.0) & \textbf{0.4065 (1)} & 0.1675 (6) & 0.2702 (5) & 0.3102 (2) & 0.2803 (3) & 0.2758 (4) & \textbf{0.4325 (1)} & 0.1286 (6) & 0.3180 (4) & 0.3141 (5) & 0.3794 (2) & 0.3397 (3) \tabularnewline
page-blocks0    & 0.7229 (3) & \textbf{0.7586 (1.0)} & 0.6823 (4) & 0.1968 (6) & 0.4094 (5) & 0.7542 (2.0) & 0.7600 (3) & 0.8123 (2) & 0.7456 (4) & 0.2099 (6) & 0.5047 (5) & \textbf{0.8248 (1) }& 0.7943 (3) & \textbf{0.8584 (1)} & 0.7534 (4) & 0.2101 (6) & 0.5066 (5) & 0.8499 (2) \tabularnewline
ecoli3          & 0.6205 (2) & 0.5905 (3.0) & 0.5540 (5) & 0.3670 (6) & 0.5748 (4) & \textbf{0.6305 (1.0)} & 0.5047 (4) & \textbf{0.5728 (1)} & 0.4843 (5) & 0.3971 (6) & 0.5471 (3) & 0.5644 (2) & 0.5464 (4) & 0.5821 (3) & 0.5441 (5) & 0.4164 (6) & 0.5914 (2) & \textbf{0.6206 (1)} \tabularnewline
yeast3          & 0.7140 (3) & 0.7225 (2.0) & 0.6569 (4) & 0.3550 (6) & 0.6568 (5) & \textbf{0.7378 (1.0)} & 0.6148 (5) & 0.6872 (2) & 0.6608 (3) & 0.4316 (6) & 0.6257 (4) & \textbf{0.7277 (1)} & 0.6855 (5) & 0.7107 (3) & 0.7175 (2) & 0.4518 (6) & 0.6897 (4) & \textbf{0.7673 (1)} \tabularnewline
new-thyroid1    & \textbf{0.8712 (1)} & 0.8147 (5.0) & 0.8635 (3) & 0.5276 (6) & 0.8268 (4) & 0.8672 (2.0) & 0.8669 (2) & 0.8228 (4) & \textbf{0.8692 (1)} & 0.6110 (6) & 0.8448 (3) & 0.8226 (5) & 0.8571 (3) & 0.8511 (4) & \textbf{0.8922 (1)} & 0.6222 (6) & 0.8608 (2) & 0.8340 (5) \tabularnewline
new-thyroid2    & 0.7783 (5) & 0.7934 (4.0) & 0.8041 (2) & 0.5129 (6) & 0.7953 (3) & \textbf{0.8154 (1.0)} & 0.8346 (2) & \textbf{0.8669 (1)} & 0.7964 (4) & 0.6337 (6) & 0.7924 (5) & 0.8325 (3) & 0.8651 (4) & \textbf{0.9003 (1)} & 0.8795 (3) & 0.6328 (6) & 0.7877 (5) & 0.8931 (2) \tabularnewline
vehicle3        & 0.5162 (4) & 0.4233 (6.0) & 0.5309 (3) & \textbf{0.5576 (1)} & 0.5099 (5) & 0.5455 (2.0) & 0.6088 (2) & 0.5411 (6) & 0.5668 (3) & \textbf{0.6097 (1)} & 0.5499 (5) & 0.5653 (4) & \textbf{0.6303 (1)} & 0.4735 (6) & 0.5870 (5) & 0.6302 (2) & 0.5890 (4) & 0.6249 (3) \tabularnewline
vehicle1        & \textbf{0.5680 (1)} & 0.4669 (6.0) & 0.5483 (5) & 0.5632 (2) & 0.5588 (3) & 0.5582 (4.0) & \textbf{0.5990 (1)} & 0.5031 (6) & 0.5469 (4) & 0.5834 (3) & 0.5205 (5) & 0.5953 (2) & \textbf{0.6161 (1)} & 0.4964 (6) & 0.6014 (4) & 0.6151 (2) & 0.5838 (5) & 0.6083 (3) \tabularnewline
vehicle2        & 0.7883 (2) & \textbf{0.8132 (1.0)} & 0.7390 (5) & 0.6918 (6) & 0.7738 (4) & 0.7852 (3.0) & \textbf{0.9095 (1)} & 0.9020 (2) & 0.8567 (5) & 0.7732 (6) & 0.8700 (4) & 0.8994 (3) & 0.9217 (2) & \textbf{0.9261 (1)} & 0.9125 (4) & 0.7849 (6) & 0.8861 (5) & 0.9142 (3) \tabularnewline
glass0          & \textbf{0.7113 (1)} & 0.6806 (4.0) & 0.6920 (3) & 0.6527 (6) & 0.6549 (5) & 0.7071 (2.0) & 0.7282 (2) & \textbf{0.7732 (1)} & 0.7087 (4) & 0.6519 (6) & 0.6533 (5) & 0.7202 (3) & 0.7331 (3) & 0.7682 (2) & \textbf{0.7776 (1)} & 0.6477 (6) & 0.6928 (5) & 0.7302 (4) \tabularnewline
pima            & \textbf{0.6527 (1) }& 0.5541 (6.0) & 0.6318 (4) & 0.6508 (2) & 0.6284 (5) & 0.6505 (3.0) & 0.4567 (2) & 0.3497 (6) & 0.3978 (4) & \textbf{0.4764 (1)} & 0.3869 (5) & 0.4476 (3) & 0.4675 (2) & 0.2823 (6) & 0.4309 (4) & \textbf{0.4735 (1)} & 0.4006 (5) & 0.4646 (3) \tabularnewline
glass1          & 0.7112 (5) & 0.7208 (3.0) & 0.7143 (4) & 0.7039 (6) & \textbf{0.7489 (1)} & 0.7298 (2.0) & \textbf{0.6650 (1)} & 0.5956 (5) & 0.6308 (3) & 0.5683 (6) & 0.6251 (4) & 0.6320 (2) & \textbf{0.6676 (1)} & 0.6617 (2) & 0.6560 (5) & 0.5856 (6) & 0.6562 (4) & 0.6579 (3) \tabularnewline
wdbc            & 0.8816 (4) & 0.8875 (3.0) & \textbf{0.8927 (1)} & 0.8775 (5) & 0.8901 (2) & 0.8756 (6.0) & \textbf{0.9118 (1)} & 0.9081 (2) & 0.8952 (4) & 0.8660 (6) & 0.8939 (5) & 0.9035 (3) & 0.9248 (4) & \textbf{0.9369 (1)} & 0.9302 (2) & 0.8898 (6) & 0.9209 (5) & 0.9260 (3) \tabularnewline
spam            & \textbf{0.7497 (1)} & 0.7391 (6.0) & 0.7406 (5) & 0.7471 (2) & 0.7436 (3) & 0.7430 (4.0) & \textbf{0.8842 (1)} & 0.8799 (2) & 0.8681 (4) & 0.8333 (6) & 0.8714 (3) & 0.8600 (5) & \textbf{0.9229 (1)} & 0.9118 (4) & 0.9194 (2) & 0.8692 (6) & 0.9158 (3) & 0.8963 (5) \tabularnewline
\hline
Avg. rank                & 2.05       & 4.17         & 3.71       & 4.76       & 3.48       & 2.83         & 2.33       & 2.95       & 3.76       & 5.00       & 4.19       & 2.76       & 2.67       & 2.95       & 3.29       & 5.14       & 4.05       & 2.90       \tabularnewline
\hline 
\end{tabular}

}
\end{table}
\end{landscape}

\end{document}